\documentclass[11pt,a4paper]{article}
\usepackage[hyperref]{acl2018}
\usepackage{times}
\usepackage{latexsym}
\usepackage{graphicx}
\usepackage{booktabs}
\usepackage{amsmath}
\usepackage{enumitem}
\usepackage{xspace}
\usepackage{relsize}
\usepackage{scrextend}
\aclfinalcopy % Uncomment this line for the final submission
\def\aclpaperid{1364} %  Enter the acl Paper ID here

\newcommand{\Custom}[1]{\textsc{#1}\xspace}

\newcommand{\model}[2][]{\textsf{#2}$_{\text{#1}}$\xspace}
\newcommand{\CNNregress}{\model{CNN$_{\text{regress}}$}}
\newcommand{\CNNregressfeat}{\model{CNN$_{\text{regress+feat}}$}}
\newcommand{\CNNregressord}{\model{CNN$_{\text{regress+ord}}$}}
\newcommand{\CNNall}{\model{CNN$_{\text{regress+ord+feat}}$}}

\newcommand{\Mean}{\model{Mean}}
\newcommand{\BoWLinear}{\model[BoW]{Linear}}
\newcommand{\GILinear}{\model[GI]{Linear}}
\newcommand{\BoWSVR}{\model[BoW]{SVR}}
\newcommand{\FeatureSVR}{\model[feat]{SVR}}
\newcommand{\BoWFeatureSVR}{\model[BoW+feat]{SVR}}

\newcommand{\secref}[2][]{Section#1~\ref{sec:#2}}
\newcommand{\tabref}[2][]{Table#1~\ref{tab:#2}}
\newcommand{\figref}[2][]{Figure#1~\ref{fig:#2}}

\title{Content-based Popularity Prediction of Online Petitions Using a Deep  Regression Model}

\author{Shivashankar Subramanian \qquad Timothy
  Baldwin \qquad Trevor Cohn  \\School of Computing and Information Systems\\ The University
  of Melbourne \\
 \url{shivashankar@student.unimelb.edu.au} \\
\qquad \url{{tbaldwin, t.cohn}@unimelb.edu.au}}

\date{}

\begin{document}
\maketitle
\begin{abstract}
  Online petitions are a cost-effective way for  citizens to collectively engage with policy-makers in a democracy. Predicting the popularity of a petition --- commonly measured by its signature count --- based on its textual content has utility for policy-makers as well as those posting the petition. In this work, we model this task using CNN regression with an auxiliary ordinal regression objective.  We  demonstrate the effectiveness of our proposed approach using UK and US government petition datasets.\footnote{
The code and data from this paper are available from  
\url{http://github.com/shivashankarrs/Petitions}}

%In this work, we address the task of predicting  popularity of petitions --- measured by the signature count, given its text. 

%In this work, we employ a CNN based deep Poisson regression model for predicting  popularity of petitions --- measured by the signature count, given its text. We also study the utility of an auxiliary ordinal-regression objective in order to handle the skew in count distribution further. Finally, we evaluate the use of custom features in addition to latent features given by the deep model. We  demonstrate the effectiveness of our proposed approach using the UK government online petitions dataset.
  
\end{abstract}

\section{Introduction}

A petition is a formal request for change or an action to any authority, co-signed by a group of supporters.  Research has shown the impact of online petitions on the political system \cite{lindner2011broadening, Hansard, bochel2017reaching}. Modeling the factors that influence petition popularity --- measured by the number of signatures a petition gets --- can provide valuable insights to policy makers as well as those authoring petitions \cite{WWW2017}. 

Previous work on modeling petition popularity has focused on predicting popularity growth over time based on an initial popularity trajectory \cite{hale2013petition, yasseri2013modeling, WWW2017}, e.g.\ given the number of signatures a petition gets in the first $x$ hours, prediction of the total number of signatures at the end of its lifetime. \citet{asherassessing} and \citet{WWW2017} examine the effect of sharing petitions on Twitter on its overall success, as a time series regression task. Other work has analyzed the importance of content on the success of the petition \cite{elnoshokaty2016success}. \citet{WWW2017} also consider the anonymity of authors and petitions featured on the front-page of the website as additional factors. \citet{huang2015activists} analyze `power' users on petition platforms, and show their influence on other petition signers. %\citet{cruickshank2009self} study whether wording suggestions 

%\footnote{\url{https://petition.parliament.uk}}
%\footnote{\url{https://petition.parliament.uk}}
In general, the target authority for a petition can be political or non-political. In this work, we use petitions from the official UK and US government websites, whereby citizens can directly appeal to the government for action on an issue. In the case of UK petitions, they are guaranteed an official response at 10k signatures, and the guarantee of parliamentary debate on the topic at 100k signatures; in the case of US petitions, they are guaranteed a response from the government at 100k signatures. Political scientists refer to this as \emph{advocacy democracy} \cite{dalton2003democracy}, in that people are able to engage with elected representatives directly. Our objective is to predict the popularity of a petition at the end of its lifetime, solely based on the petition text.  

\citet{elnoshokaty2016success} is the closest work to this paper, whereby they target Change.org petitions and perform correlation analysis of popularity with the petition's category, target goal set,\footnote{See \url{http://bit.ly/2BXd0Sl}.} and the distribution of words in General Inquirer categories \cite{stone1962general}. In our case, we are interested in the task of automatically predicting the number of signatures.

%More importantly, we handle the skewness or long-tail observed in the signature count data \cite{hale2013petition, yasseri2013modeling} using a deep Poisson regression model \cite{mccullagh1984generalized}. 
We build on the convolutional neural network (CNN) text regression model of \citet{bitvai2015non} to infer deep latent features. In addition, we evaluate the effect of an auxiliary ordinal regression objective, which can discriminate petitions that attract different scales of popularity (e.g., 10 signatures, the minimum count needed to not be closed vs.\ 10k signatures, the minimum count to receive a response from UK government).

Finally, motivated by text-based message propagation analysis work \cite{tan2014effect, piotrkowicz2017headlines}, we hand-engineer features which capture wording effects on petition popularity, and measure the ability of the deep model to automatically infer those features.

\section{Proposed Approach}
\label{sec:approach}

\begin{figure}
\centering
\includegraphics[scale=0.27]{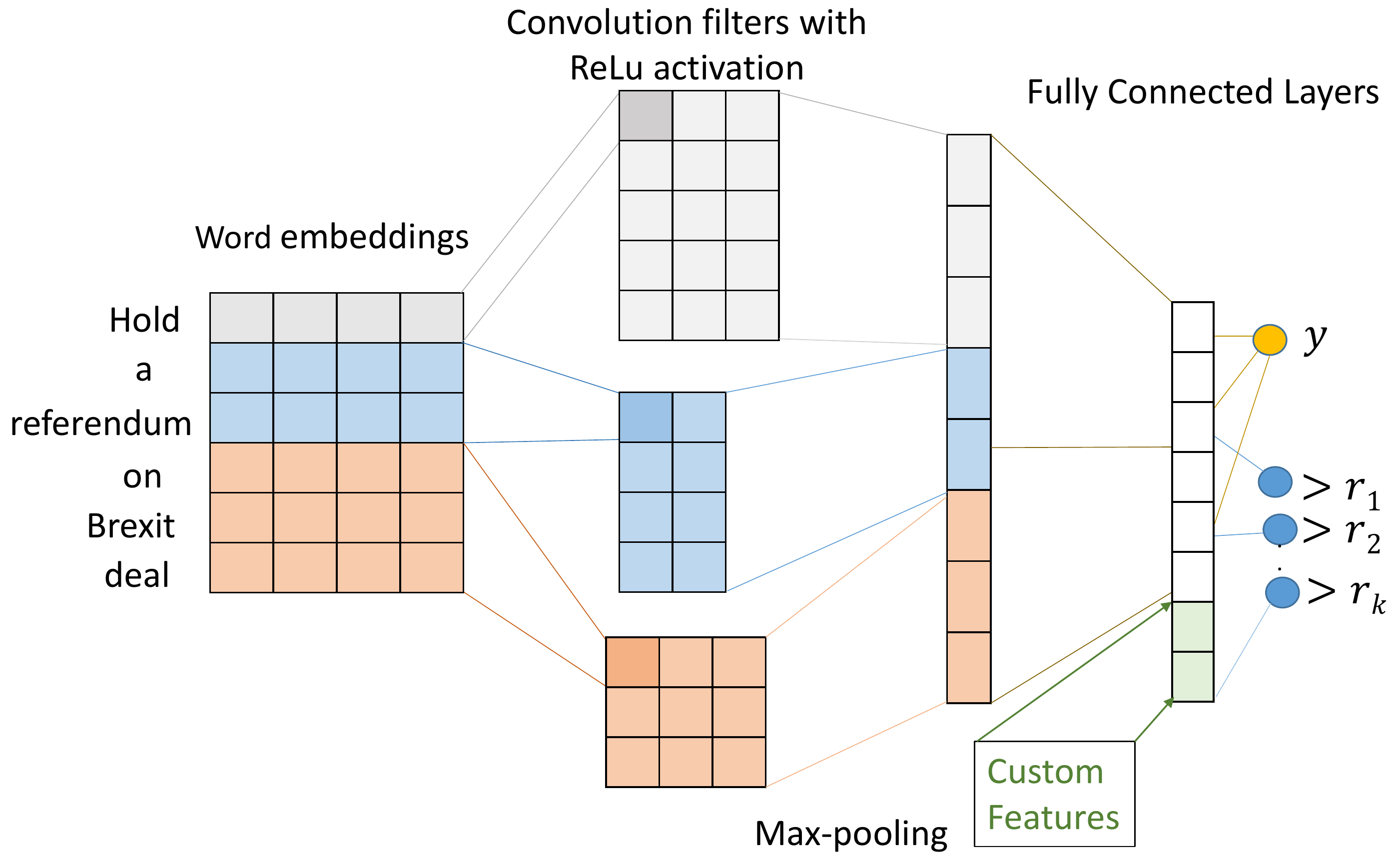}
\caption{CNN-Regression Model. $y$ denotes signature count. $>r_{k}$ is the auxiliary task that denotes $p$(petition attracting $>r_{k}$ signatures).}
\label{fig:CNN}
\end{figure}

Inspired by the successes of CNN for text categorization \cite{kim2014convolutional} and text regression \cite{bitvai2015non}, we propose a CNN-based model for predicting the signature count. An outline of the model is provided in \figref{CNN}. A petition has three parts: (1) title, (2) main content, and (3) (optionally) additional details.\footnote{Applicable for the UK government petitions only.} We concatenate all three parts to form a single document for each petition. We have $n$ petitions as input training examples of the form \{$a_{i}$, $y_{i}$\}, where $a_{i}$ and $y_{i}$ denote the text and signature count of petition $i$, respectively. Note that we log-transform the signature count, consistent with previous work \cite{elnoshokaty2016success,WWW2017}.

We represent each token in the document via its pretrained GloVe embedding \cite{pennington2014glove}, which we update during learning. We then apply multiple convolution filters with width one, two and three to the dense input document matrix, and apply a ReLU to each. They are then passed through a max-pooling layer with a $\tanh$ activation function, and finally a multi-layer perceptron via the exponential linear unit activation,
\[
    f(x)= 
\begin{cases}
    x, & \text{if } x > 0 \\
    \alpha\left(\exp(x)-1\right)              & \text{otherwise}\,,
\end{cases}
\]
to obtain the final output ($y_{i}$), which is guaranteed to be positive. 
We train the model by minimizing mean squared error in log-space,

\begin{equation}
\mathcal{L}_{reg} = \frac{1}{n}  \sum_{i=1}^{n} \|\hat{y}_i - y_i\|^2_2 \,,
\label{eq:doc-loss}
\end{equation}
where $\hat{y}_i$ is the estimated signature count for petition $i$. We refer to this model as \CNNregress.

\subsection{Auxiliary Ordinal Regression Task}
We augment the regression objective with an ordinal regression task, which discriminates petitions that achieve different scale of signatures. The intuition behind this is that there are pre-determined thresholds on signatures which trigger different events, with the most important of these being 10k (to guarantee a government response) and 100k (to trigger a parliamentary debate) for the UK petitions; and 100k (to get a government response) for the US petitions. In addition to predicting the number of signatures, we would like to be able to predict whether a petition is likely to meet these thresholds, and to this end we use the exponential ordinal scale based on the thresholds $O = \{10,100,1000,10000,100000\}$.\footnote{We use $O = \{1000,10000,100000\}$ for the US petitions, as only petitions which get a minimum of 150 signatures are published on the website.} Overall this follows the exponential distribution of signature counts closely \cite{yasseri2013modeling}.

We transform the ordinal regression problem into a series of simpler binary classification subproblems, as proposed by \citet{li2007ordinal}. We construct binary classification objectives for each threshold in $O$. For each petition $i$ we construct an additional binary vector $\vec{o}_i$, with a 0--1 encoding for each of the ordinal classes (\{$a_{i}$,$y_{i}$,$\vec{o}_i$\}). Note that the transformation is done in a consistent way, i.e., if a petition has $y$ signatures, then in addition to immediate lower-bound threshold in $O$ determined by $l = \lfloor\log_{10} y\rfloor$ (for $y < 10^6$), all classes which have a lesser threshold are also set to 1 ($o_{t:  t<l}$).

With this transformation, apart from the real-valued output $y_i$, we also learn a mapping from $\mathbf{h}_{i}$ with sigmoid activation for each class ($\vec{r}_{i}$). Finally we minimize cross-entropy loss for each binary classification task, denoted $\mathcal{L}_{aux}$.

% \begin{equation}
% \mathcal{L}_{aux} = \frac{1}{n} \sum_{i=1}^{n} \text{cross-entropy}(\vec{r_{i}}, \vec{o_{i}}) 
% \label{eq:aux}
% \end{equation}

Overall, the loss function for the joint model is:
\begin{equation}
 \mathcal{L}_J = \mathcal{L}_{reg} + \gamma \mathcal{L}_{aux}
 \label{eq:joint-loss}
\end{equation}
where $\gamma \ge 0$ is a hyper-parameter which is tuned on the validation set. We refer to this model as \CNNregressord.

%It has the test statistic \emph{H}, and the $p$-value is computed with \emph{H} following a chi-squared distribution.

\section{Hand-engineered Features}
\label{sec:features}

We hand-engineered custom features, partly based on previous work on non-petition text. This includes features from \newcite{tan2014effect} and  \newcite{piotrkowicz2017headlines} such as structure, syntax, bias, polarity, informativeness of title, and novelty (or freshness), in addition to novel features developed specifically for our task, such as policy category and political bias features. We provide a brief description of the features below:
%, and provide further details in the supplementary material
\begin{itemize}[nosep,leftmargin=1em,labelwidth=*,align=left]
\item Additional Information (\Custom{Add}): binary flag indicating whether the petition has additional details or not.
%\footnote{Applicable for the UK government petitions only.}
% Lookup features ,***, $+$
\item Ratio of indefinite (\Custom{Ind}) and definite (\Custom{Def}) articles. %,*, $-$ ,***, $+$
\item Ratio of first-person singular pronouns (\Custom{Fsp}), first-person plural pronouns (\Custom{Fpp}), second-person pronouns (\Custom{Spp}), third-person singular pronouns (\Custom{Tsp}), and third-person plural pronouns (\Custom{Tpp}). %,***, $-$ ,*, $+$ ,*, $-$ ,-- ,**, $-$
\item Ratio of subjective words (\Custom{Subj}) and difference between count of positive and negative words (\Custom{Pol}), based on General Inquirer lexicon. % ,*, $-$ ,--
\item Ratio of biased words (\Custom{Bias}) from the bias lexicon \cite{recasens2013linguistic}. % ,*, $-$
\item Syntactic features: number of nouns (\Custom{NNC}), verbs (\Custom{VBC}), adjectives (\Custom{ADC}) and adverbs (\Custom{RBC}). % ,-- ,-- ,***, $-$ ,**, $-$
\item Number of named entities (\Custom{NEC}), based on the NLTK NER model \cite{Bird+:2009}. % ,***, $+$
\item Freshness (\Custom{Fre}): cosine similarity with all previous petitions, inverse weighted by the difference in start date of petitions (in weeks). %,***, $-$ \footnote{See \url{https://petition.parliament.uk/help}}
\item Action score of title (\Custom{Act}): probability of title conveying the action requested. Predictions are obtained using an one-class SVM model built on the universal representation \cite{conneau2017supervised} of titles of rejected petitions,\footnote{\url{https://petition.parliament.uk/help}} as they don't contain any action request. These rejected petitions are not part of our evaluation dataset.  
% , --  (Section 4)
\item Policy category popularity score (\Custom{Csc}): commonality of the petition's policy issue \cite{ALTW2017}, based on the recent UK/US election manifesto promises. %,***, $+$
\item Political bias and polarity: relative leaning/polarity based on:  (a) $\frac{\#\text{left} + \#\text{right}}{\#\text{left} + \#\text{right} + \#\text{neutral}}$ (\Custom{Pbias}) (b) $\frac{\#\text{left} - \#\text{right}}{\#\text{left} + \#\text{right}}$ (\Custom{L--R}). Sentence-level \emph{left}, \emph{right} and \emph{neutral} classes are obtained using a model built on the CMP dataset, and the categorization given by \citet{budge2013}.
\end{itemize}

The custom features are passed through a hidden layer with $\tanh$ activations ($\mathbf{c}_{i}$), and concatenated with the hidden representation learnt using the dense input document (\secref{approach}), $\left[\begin{array}{cc} \mathbf{h}_{i} \\ \mathbf{c}_{i} \end{array} \right]$, before  mapping to the output layer (\figref{CNN}). We refer to this model as \CNNall. We use the Adam optimizer \cite{KingmaB14} to train all our models.

\section{Evaluation}
\begin{figure}[t]
%\begin{subfigure}{0.56\textwidth}
\centering
\includegraphics[width=0.51\linewidth]{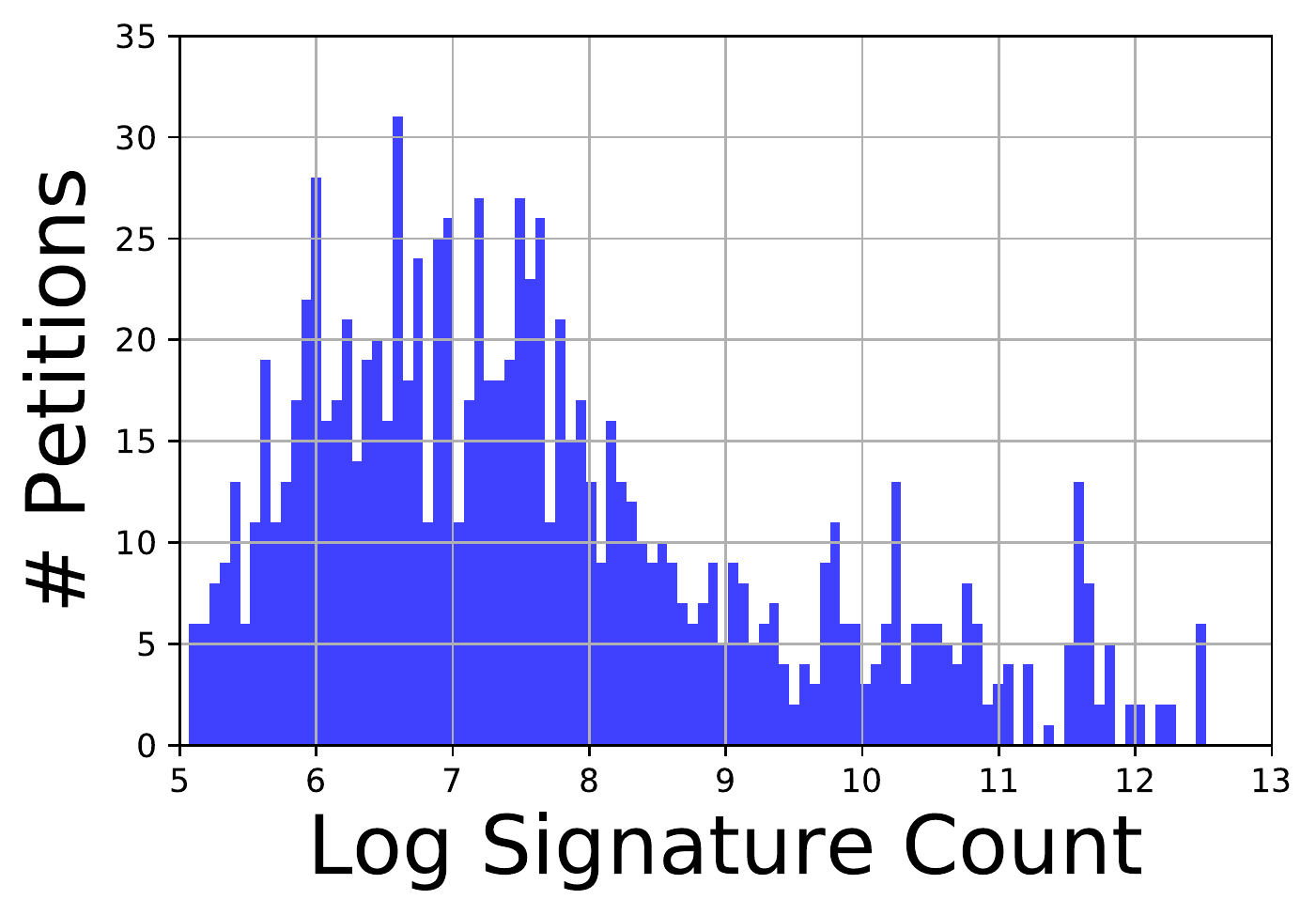}
\vspace{-1ex}
\caption{US Petitions Signature Distribution}
\label{fig:US}
%\end{subfigure}
\end{figure}

%\footnote{\url{http://reshare.ukdataservice.ac.uk/851634/}} 
We collected our data from the UK\footnote{\url{https://petition.parliament.uk}} and US\footnote{\url{https://petitions.whitehouse.gov/}} government websites
over the term of the 2015--17 Conservative and 2011--14 Democratic governments respectively. The UK dataset contains 10950 published petitions, with over 31m signatures in total. We removed US petitions with $\le 150$ signatures, resulting in a total of 1023 petitions, with over 12m signatures in total. We split the data chronologically into train/dev/test splits based on a 80/10/10 breakdown. Distribution over $\log$ signature counts is given in \figref[s]{UK} and \ref{fig:US}. 

\begin{figure}[t]
%\begin{subfigure}{0.56\textwidth}
\centering
\includegraphics[width=0.51\linewidth]{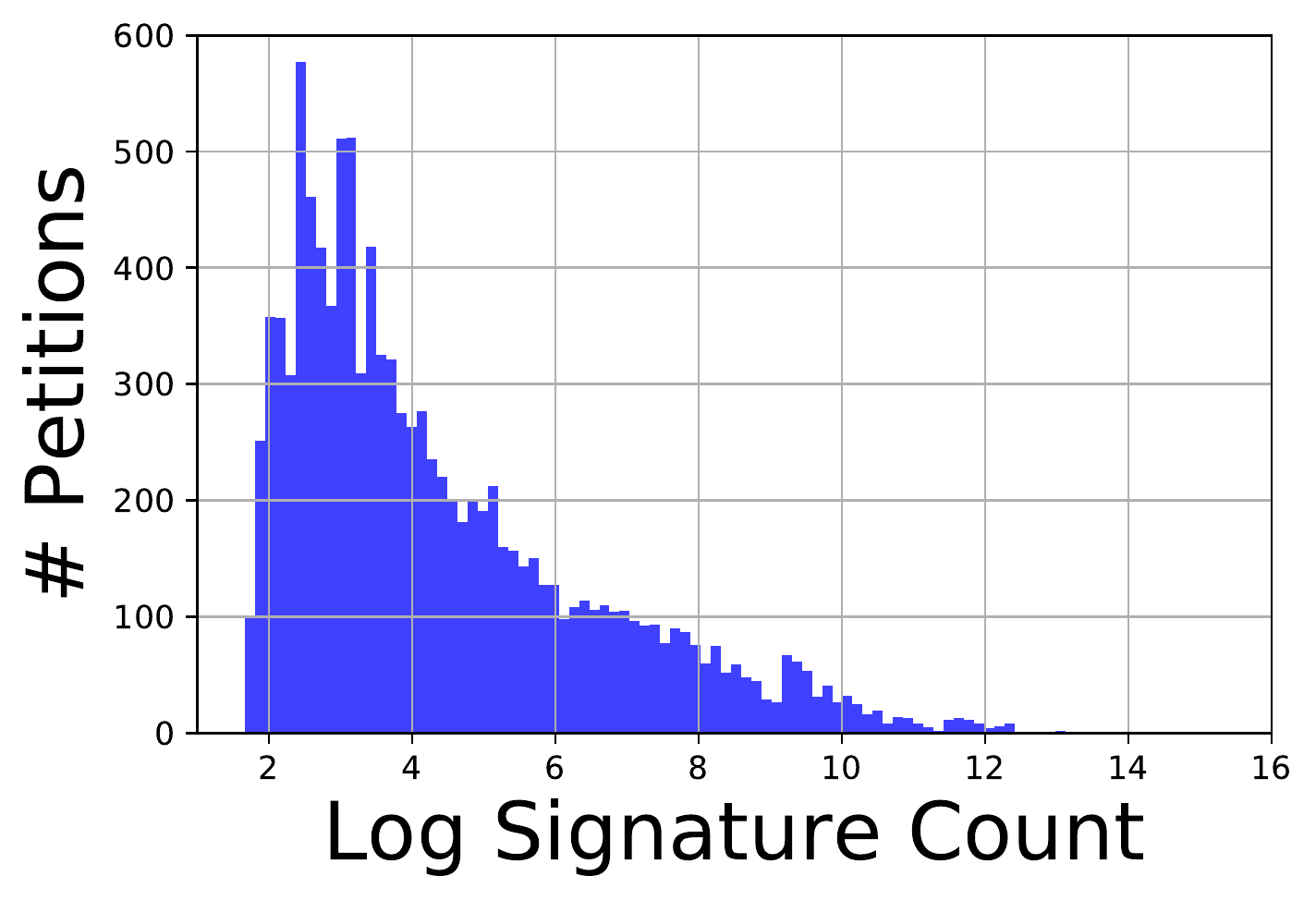}
\vspace{-1ex}
\caption{UK Petitions Signature Distribution.}
\label{fig:UK}
\end{figure}
%\end{subfigure} \hspace{0 \textwidth}

To analyze the statistical significance of each feature varying across ordinal groups $O$, we ran a Kruskal-Wallis test (at $\alpha=0.05$: \newcite{kruskal1952use}) on the training set. The test results in the test statistic $H$ and the corresponding $p$-value, with a high $H$ indicating that there is a difference between the two groups. The analysis is given in \tabref{clra}, where $p < 0.001$, $p < 0.01$ and $p < 0.05$ are denoted as ``***'', ``**'' and ``*'', respectively. Note that the ordinal groups are different for the two datasets: analyzing the UK dataset with the same ordinal groups used for the US dataset (\{1000,10000,100000\}) resulted in a similarly sparse set of significance values for non-syntactic features as the US dataset.
%$p$-value is computed assuming that \emph{H} follows a chi-squared distribution. 
%$p$-value is computed assuming that \emph{H} follows a chi-squared distribution.
%For features where there was statistical significance ($p < 0.05$), we also provide the linear correlation (intercept value, $\beta$, of a linear model with each custom feature as independent variable and signature count as dependent variable) to interpret its effect on popularity, e.g., having additional details (\Custom{Add}) has a positive influence on the signature count. 

We benchmark our proposed approach against the following baseline approaches:
%Kruskal-Wallis-H test (KWH) \cite{kruskal1952use} to analyze the statistical significance of each custom feature to vary across ordinal groups $O$ using the training set is given Table \ref{tab:KWH}. It has the test statistic (\emph{H}), which when high indicates that there is a difference between two or more groups, and the corresponding p-value (\emph{p}). Note that only significant results ($p<$0.05) are included. We also provide the sign of correlation, to understand its effect on the popularity, e.g., having additional details (Add) has a positive correlation.  We compare against the following baseline approaches:
%\begin{description}[noitemsep]
\begin{description}[nosep,leftmargin=1em,labelwidth=*,align=left]
\item[\Mean:] average signature count in the raining set.
\item[\BoWLinear:] linear regression (Linear) model using TF-IDF weighted bag-of-words features.
\item[\GILinear:] linear regression model based on word distributions from the General Inquirer lexicon; similar to \cite{elnoshokaty2016success}, but without the target goal set or category of the petition (neither of which is relevant to our datasets).
\item[\BoWSVR:] support vector regression (SVR) model with RBF kernel and TF-IDF weighted bag-of-words features.
\item[\FeatureSVR:] SVR model using the hand-engineered features from \secref{features}.
\item[\BoWFeatureSVR:] SVR model using combined TF-IDF weighted bag-of-words and hand-engineered features.
%\item[SVMord]: SVM-based ordinal classification over the three most significant classes of $[0,10000)$, $[10000,100000)$ and $[100000,\infty)$ \cite{li2007ordinal}.
\end{description}

%  \begin{table*}[!ht]
%   \centering
%   \begin{small}
%   \begin{tabular}{ l c c }
%   \toprule
%     Approach & MAE & MAPE \\
%     \midrule
%     Mean    &  4.37 & 159.7  \\
%     BoW LR    &  1.75 & 57.56  \\
%     GI LR &  1.77 & 58.22  \\
%     BoW SVR    &  1.53 & 45.35  \\    
%     Custom SVR &  1.54 & 46.96  \\    
%     BoW+Custom SVR    &  1.52 & 44.71  \\    
%     %Custom LR &  1.52 & 998.5 & \\
%     %GI LR &  1.77 & & \\
%     %Custom SVR &  1.53 & 977 & \\
%     %BoW+Custom LR &  1.57 & & \\
%     %BoW+Custom SVR &  1.52 & & \\
%     \midrule
%     CNN$_{regress}$ &  1.44 & 36.72\\
%     CNN$_{regress+ord}$ &  1.42 & 33.86 \\
%     CNN$_{regress+ord+cust}$ &  1.41 & 32.92 \\
%     CNN$_{regress+cust}$ &  1.43 & 35.84 \\
%     CNN$_{regress+ord+cust}$ + Additional hidden layer &  \textbf{1.40} & \textbf{31.68}\\
%     \bottomrule
%   \end{tabular}
%   \end{small}
%   \caption{Test-set error. Proposed approach variants are given in last block. Best scores are given in bold.}
%  \label{tab:mae}
% \end{table*}

% SS We can comment out the below table and use the above one.

 \begin{table*}[!t]
  \centering
  \begin{small}
  \begin{tabular}{ lc c c@{\;} c c }
%  \toprule
   &  \multicolumn{2}{c}{UK Petitions} && \multicolumn{2}{c}{US Petitions} \\
    \cmidrule{2-3}
    \cmidrule{5-6}
  Approach & MAE & MAPE && MAE & MAPE \\
    %Approach & MAE & MAPE & F-score\\
    \toprule
    \Mean    &  4.37 & 159.7 && 2.82 & 44.61 \\
    \BoWLinear    &  1.75 & 57.56 && 2.51 & 37.01 \\
    \GILinear &  1.77 & 58.22 && 1.84 & 27.71 \\
    \BoWSVR    &  1.53 & 45.35 && 1.39 & 20.37 \\    
    \FeatureSVR &  1.54 & 46.96 && 1.40 & 20.48 \\    
    \BoWFeatureSVR    &  1.52 & 44.71 && 1.39 & 20.38 \\    
    %\SVMord    &  --- & --- && 0.33 \\    
    %Custom LR &  1.52 & 998.5 & \\
    %GI LR &  1.77 & & \\
    %Custom SVR &  1.53 & 977 & \\
    %BoW+Custom LR &  1.57 & & \\
    %BoW+Custom SVR &  1.52 & & \\
    \midrule
    \CNNregress &  1.44 & 36.72 && 1.24 & 14.98\\
    \CNNregressord &  1.42 & 33.86 && 1.22 & 14.68 \\
    \CNNall &  1.41 & 32.92 && 1.20 & 14.47 \\
    \CNNregressfeat &  1.43 & 35.84 && 1.23 & 14.75\\
    \CNNall + Additional hidden layer &  \textbf{1.40} & \textbf{31.68} && \textbf{1.16}  & \textbf{14.38}\\
    \bottomrule
  \end{tabular}
  \end{small}
  \caption{Results over UK and US Government petition datasets. Best scores are given in bold.}
 \label{tab:mae}
\end{table*}

We present the regression results for the baseline and proposed approaches based on: (1) mean absolute error (MAE), and (2) mean absolute percentage error (MAPE, similar to \citet{WWW2017}), calculated as
%and macro-averaged F-score in \tabref{mae}. MAPE is a relative percentage measure:
$\frac{100}{n}\sum_{i=1}^{n}\frac{|\hat{y_i}-y_i|}{y_i}$. Results are given in \tabref{mae}.

The proposed CNN models outperform all of the baselines. Comparing the CNN model with regression loss only, \CNNregress, and the joint model, \CNNregressord is superior across both datasets and measures. When we add the hand-engineered features (\CNNall), there is a very small improvement. In order to further understand the effect of the hand-engineering features without the ordinal regression loss, we use it only with the regression task (\CNNregressfeat), which mildly improves over \CNNregress, but is below \CNNall.  We also evaluate a variant of \CNNall with an additional hidden layer, given in the final row of \tabref{mae}, and find it to lead to further improvements in the regression results. Adding more hidden layers did not show further improvements.

%The F-score is calculated over the three classes of $[0,10000)$, $[10000,100000)$ and $[100000,\infty)$ (corresponding to the thresholds at which the petition leads to a government response or parliamentary debate), by determining if the predicted and actual signature counts are in the same bin or not.
\subsection{Classification Performance}
The F-score is calculated over the three classes of $[0,10000)$, $[10000,100000)$ and $[100000,\infty)$ (corresponding to the thresholds at which the petition leads to a government response or parliamentary debate) for the UK dataset; and $[150,100000)$ and $[100000,\infty)$ for the US dataset, by determining if the predicted and actual signature counts are in the same bin or not. We also built an SVM-based ordinal classifier \cite{li2007ordinal} over the significant ordinal classes, as an additional baseline. The CNN models struggle to improve F-score (in large part due to the imbalanced data). For the UK dataset, CNN models with an ordinal objective (\CNNregressord and \CNNall) result in a macro-averaged F-score of 0.36, compared to 0.33 for all other methods. But for the US dataset, which is a binary classification task, all methods obtain a 0.49 F-score. In addition to text, considering other factors such as early signature growth \cite{hale2013petition} --- which determines the timeliness to get the issue online on the US website --- could be necessary. 
%--- over 2\% of the test data falls in the $[10000,100000)$ and $[100000,\infty)$ bins put together for the UK dataset

\setlength{\tabcolsep}{1em}
\begin{table*}[!t]
  \centering
  \begin{small}
  \begin{tabular}{ll r c c c@{\;} r c c}
%  \toprule
   & &  \multicolumn{3}{c}{UK Petitions} && \multicolumn{3}{c}{US Petitions} \\
    \cmidrule{3-5}
    \cmidrule{7-9}
    
  Feature & Description & \multicolumn{1}{c}{$H$} & $p$ & $p_{\text{hidden}}$ && \multicolumn{1}{c}{$H$} & $p$ & $p_{\text{hidden}}$ \\
  \toprule
  %\footnote{US government site does not have additional details}
  \Custom{Add} & Additional details & 94.59  &  *** &  && 	&   & \\
    \Custom{Ind} & Indefinite articles  & 14.87 &  *  &  && 8.56 & * &   *
	\\
     \Custom{Def} & Definite articles  & 34.91 & ***  & * && 3.69 &  & \\    	  
      \Custom{Fsp} & First-person singular pronouns & 53.36 &  ***  &  && 6.84 & *  & \\
     \Custom{Fpp}  & First-person plural pronouns  &11.26 & * &  && 6.10 &  & \\    
      \Custom{Spp}  & Second-person pronouns  & 13.80 & *  &  && 3.95  &  & \\   
    \Custom{Tsp}  & Third-person singular pronouns  & 5.82 &  &  && 9.07 & *  & \\
    \Custom{Tpp}  & Third-person plural pronouns  & 16.13 & **&  && 5.58 &  & \\
    \Custom{Subj}  & Subjective words  & 12.25 & * &  && 7.21 & *  & ***\\
    \Custom{Pol}  & Polarity  & 2.60 &   & * && 4.27 &  & \\
    \Custom{Bias} & Biased words   & 11.92 & * &  && 4.56 & *  & \\            
    \Custom{NNC} & Nouns& 7.34 &   & *** && 1.93 &  & **\\
    \Custom{VBC} & Verbs & 2.75 &   & ** && 7.46 & *  & ***\\
    \Custom{ADC} & Adjectives & 26.14 & ***  & *** && 4.07 &  & \\
    \Custom{RBC} & Adverbs & 17.09& **  &  && 2.99 &  & *\\    
    \Custom{NEC}    & Named entities & 51.11  & ***  &  *** && 3.94  &  & *\\
    \Custom{Fre}    & Freshness & 86.97 & ***  & * && 13.86 & ** & *\\
    \Custom{Act}    & Title's action score & 3.89 &  &  && 3.54 &  & \\
    \Custom{Csc}    & Policy category popularity & 38.22 & *** &  &&1.94 &  & \\
    \Custom{Pbias}    & Political bias & 4.13 &  &  && 12.23 & ** & \\
    \Custom{L--R}    & Left--right scale & 10.94 & *  &  && 12.88 & ** & \\  
  \bottomrule
  \end{tabular}
\end{small}
\caption{Dependency of hand-engineered features against the signature count ($p$ and $H$) and deep hidden features ($p_{\text{hidden}}$). \Custom{Add} is not applicable for the US government petitions dataset. $p < 0.001$, $p < 0.01$ and $p < 0.05$ are denoted as ``***'', ``**'' and ``*'', respectively.}
\label{tab:clra}
\end{table*}

\subsection{Latent vs.\ Hand-engineered Features}

Finally, we built a linear regression model with the estimated hidden features from  \CNNregressord as independent variables and hand-engineered features as dependent variables, to study their linear dependencies in a pair-wise fashion. The most significant dependencies (given by $p$-value, $p_{\text{hidden}}$) over the test set are given in \tabref{clra}. We found that the model is able to learn latent feature representations for syntactic features (\Custom{NNC}, \Custom{VBC}, $\Custom{ADC}$,\footref{foot:ukonly} $\Custom{RBC}$\footref{foot:usonly}), \Custom{Fre}, \Custom{NEC},  \Custom{Ind} and $\Custom{Def}$,\footref{foot:ukonly} but not the other features --- these can be considered to provide deeper information than can be extracted automatically from the data, or else information that has no utility for the signature prediction task.  From the analysis in \tabref{clra}, some of the features that vary across ordinal groups are not linearly dependent with the deep latent features. These include $\Custom{ADD}$,\footref{foot:ukonly} \Custom{Bias}, \Custom{Csc},\footnote{\label{foot:ukonly}UK dataset only.} \Custom{Pbias},\footnote{\label{foot:usonly}US dataset only.} and \Custom{L--R}, where the latter ones are policy-related features. This indicates that the custom features and hidden features contain complementary signals.

Overall our proposed approach with the auxiliary loss and hand-engineered features (\CNNall) provides a reduction in MAE over \CNNregress by 2.1\% and 3.2\%, and SVR by 7.2\% and 13.7\% on the UK and US datasets, resp. Although the ordinal classification performance is not very high, it must be noted that the data is heavily skewed (only 2\% of the UK test-set falls in the $[10000,100000)$ and $[100000,\infty)$ bins put together), and we tuned the hyper-parameters wrt the regression task only.

\section{Conclusion and Future Work}

This paper has targeted the prediction of the popularity of petitions directed at the UK and US governments. In addition to introducing a novel task and dataset, contributions of our work include: (a) we have shown the utility of an auxiliary ordinal regression objective; and (b) determined which hand-engineered features are complementary to our deep learning model. In the future, we aim to study other factors that can influence petition popularity in conjunction with text, e.g., social media campaigns, news coverage, and early growth rates. 
%Also building a global model for different countries using all the text together.
%, news coverage, etc.
%The joint model with an auxiliary objective and custom features together outperforms other approaches.

\section*{Acknowledgements}
We thank the reviewers for their valuable
comments. This
work was funded in part by the Australian Government
Research Training Program Scholarship, and
the Australian Research Council. 

\bibliography{acl2018}
\bibliographystyle{acl_natbib}

\end{document}